\pdfoutput=1

\documentclass[11pt]{article}

\usepackage[a-1b]{pdfx} %

\usepackage{acl}

\usepackage{times}
\usepackage{latexsym}

\usepackage[T1]{fontenc}

\usepackage[utf8]{inputenc}

\usepackage{microtype}

\usepackage[a-1b]{pdfx}
\usepackage{hyperref}
\usepackage[utf8]{inputenc}
\usepackage{colorprofiles}
\hypersetup{pdfstartview=}  %

\usepackage{amsmath}
\usepackage{amssymb}
\usepackage{amsfonts}
\usepackage{amsthm}
\usepackage{booktabs}
\usepackage{graphicx}
\usepackage{multirow}
\usepackage{subcaption}
\usepackage{tabularx}
\usepackage{comment}

\usepackage{amsmath,amsfonts,bm}

\def\eqref#1{equation~\ref{#1}}

\def\1{\bm{1}}

\def\ve{{\mathbf{e}}}

\def\vk{{\mathbf{k}}}

\def\vq{{\mathbf{q}}}

\def\vv{{\mathbf{v}}}
\def\vw{{\mathbf{w}}}
\def\vx{{\mathbf{x}}}
\def\vy{{\mathbf{y}}}
\def\vz{{\mathbf{z}}}
\def\vphi{{\boldsymbol{\phi}}}

\def\mS{{\mathbf{S}}}

\DeclareMathAlphabet{\mathsfit}{\encodingdefault}{\sfdefault}{m}{sl}
\SetMathAlphabet{\mathsfit}{bold}{\encodingdefault}{\sfdefault}{bx}{n}

\definecolor{orange}{rgb}{1,0.5,0}
\definecolor{mdgreen}{rgb}{0.05,0.6,0.05}
\definecolor{mdblue}{rgb}{0,0,0.7}
\definecolor{dkblue}{rgb}{0,0,0.5}
\definecolor{dkgray}{rgb}{0.3,0.3,0.3}
\definecolor{slate}{rgb}{0.25,0.25,0.4}
\definecolor{gray}{rgb}{0.5,0.5,0.5}
\definecolor{ltgray}{rgb}{0.7,0.7,0.7}
\definecolor{purple}{rgb}{0.7,0,1.0}
\definecolor{lavender}{rgb}{0.65,0.55,1.0}

\newcommand{\ensuretext}[1]{#1}

\newcommand{\arkcomment}[3]{\ensuretext{\textcolor{#3}{[#1 #2]}}}
\renewcommand{\arkcomment}[3]{}  %

\definecolor{revcolor}{rgb}{0.05,0.8,0.05}
\newcommand{\rev}[1]{\textcolor{black}{#1}}

\title{Modeling Context With Linear Attention for \\ Scalable Document-Level Translation} %

\usepackage{wasysym}
\usepackage{marvosym}
\author{Zhaofeng Wu$^\text{\Cancer}$ \quad
    Hao Peng$^\text{\Leo}$ \quad
    Nikolaos Pappas$^\text{\Gemini}$ \quad
    Noah A. Smith$^{\text{\Leo\ } \rotatebox[y=0.1cm]{90}{\textsuperscript{\Cancer}}}$ \\
    $^\text{\Cancer}$MIT \quad $^\text{\Leo}$Allen Institute for Artificial Intelligence \quad $^\text{\Gemini}$AWS AI \\
    $^{\rotatebox[y=0.1cm]{90}{\textsuperscript{\Cancer}}}$\hspace{-0.1cm}Paul G. Allen School of Computer Science \& Engineering, University of Washington \\
    \texttt{zfw@csail.mit.edu \quad haop@allenai.org} \\
    \texttt{nppappa@amazon.com \quad nasmith@cs.washington.edu}
}

\begin{document}
\maketitle

\begin{abstract}
Document-level machine translation leverages inter-sentence dependencies to produce more coherent and consistent translations. However, these models, predominantly based on transformers, are difficult to scale to long documents as their attention layers have quadratic complexity in the sequence length. 
Recent efforts on efficient attention improve scalability, but their effect on document translation remains unexplored.
In this work, we investigate the efficacy of a recent linear attention model by \citet{peng2021rfa} on document translation and augment it with a sentential gate to promote a recency inductive bias. 
We evaluate the model on IWSLT 2015 and OpenSubtitles 2018 against the transformer, \rev{demonstrating substantially increased decoding speed on long sequences} with similar or better BLEU scores. We show that sentential gating further improves translation quality on IWSLT.\textsuperscript{\ref{fn:code}}%

\end{abstract} %

\section{Introduction} \label{sec:intro}
Sentence-level neural machine translation has seen significant recent progress~\cite{Bahdanau2015, vaswani2017}. A move to document-level 
translation opens the possibility of using inter-sentential context at the scale of paragraphs, documents, or even whole books~\cite{lopes-etal-2020-document,ma2021comparison,maruf2021survey}. This opens up new research
avenues to improve translation and its evaluation for more consistent anaphora resolution and 
discourse coherence~\cite{bawden-etal-2018-evaluating,muller-etal-2018-large,voita-etal-2019-good}.

Transformers have enabled state-of-the-art results for sentence-level translation~\cite{vaswani2017,chen-etal-2018-best,wang-etal-2019-learning} and this success has made them the default architecture for document translation.
However, they do not scale well in the sequence length due to the quadratic complexity of attention
and hence are computationally prohibitive to apply to long text. %
Alternative architectures exist, but
most still have quadratic complexity~\cite{zhang-etal-2018-improving,voita-etal-2019-good} and/or have extra modules that further add to the inference cost \cite{tu2017learning,zhang-etal-2018-improving,miculicich-etal-2018-document,donato-etal-2021-diverse}.

By reducing asymptotic complexity, recent work on efficient attention may pave the way for long text generation.
However, these methods' suitability for document translation requires further investigation:
some do not focus on decoding speed~\citep[\emph{i.a.}]{guo-etal-2019-star,child19,Kitaev2020Reformer,wang2020linformer}, while others' speedup and quality impact for document translation remains unknown~\citep[\emph{i.a.}]{kasai2021finetuning,schlag2021linear,ma2021luna,choromanski2021rethinking}. %
In this work, we consider random feature attention (RFA; \citealp{peng2021rfa}), a representative first model with established accuracy and efficiency in sentence-level translation.
With few additional parameters, it approximates softmax attention in linear time and space with recurrent computations.
We explore its effectiveness for document translation %
and find substantial decoding speedup over a transformer with similar or improved BLEU. 
We also equip RFA with a sentential gate, injecting a recency inductive bias tailored to representing document context.%

Our main contributions are: (i) we study the efficacy of RFA for document translation; 
(ii) we validate that RFA is competitive with a transformer \rev{but substantially faster} on long documents;
(iii) we augment RFA with a sentential gate designed to promote a recency bias, which brings
a 0.5 BLEU improvement on IWSLT~\cite{Cettolo2015TheI2}. 
To encourage future research, we  release our code.\footnote{\url{https://github.com/ZhaofengWu/rfa-doc-mt}\label{fn:code}}

\section{Background} \label{sec:doc-mt}

\begin{figure}[t]
\centering
\includegraphics[width=0.46\textwidth]{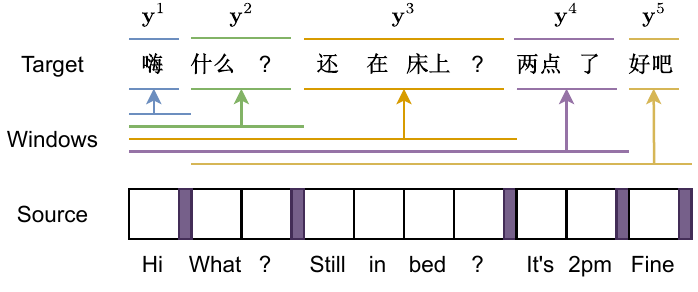}
\caption{\label{fig:sliding-window} The concatenation model for document translation with a sliding window of length $L=4$. Every window is translated in its entirety, but only the last translated sentence is used for evaluation. The purple bars denote the sentence separator token.
}
\end{figure}

Standard machine translation independently translates each sentence.
Document translation leverages additional source and target context to produce more coherent translation, improving lexical choice and ambiguity resolution~\cite{voita-etal-2019-good}.

\paragraph{The Concatenation Model.}
Recent studies found that the concatenation model that directly translates the source document
to the target document (or a multi-sentence window) with a single encoder-decoder model performs well~\cite{tiedemann-scherrer-2017-neural,ma2021comparison}, especially on large datasets~\cite{junczysdowmunt2019microsoft}. Figure~\ref{fig:sliding-window} illustrates this model combined with sliding window decoding, which we adopt in this work.

\paragraph{Scalability of Attention.} 
In machine translation, transformers have three types of attention: encoder self-attention, cross attention, and causal attention. %
In each, every query $\vq_t$ is dotted with all keys $\{\vk_i\}$ to obtain the attention weights, with which a weighted average of the values $\{\vv_i\}$ is calculated:
\begin{align}\label{eq:attention}
\begin{split}
  \hspace{-1mm}\operatorname{attn}\left(\vq_t,\{\vk_i\},\{\vv_i\}\right)=\sum_{i=1}^{N}
    \frac{\exp\left(\vq_t \cdot \vk_i \right)}
    {\sum_{j=1}^{N}\exp\left(\vq_t \cdot \vk_{j} \right)}\vv_i^\top \nonumber
\end{split}
\end{align}
\noindent where $N$ is the sequence length. This pairwise interaction incurs quadratic overhead
in $N$, which is inefficient for the long text sequences in the concatenation model. 
This particularly impacts cross and causal attention at decoding time, which cannot be parallelized~\citep{kasai2021finetuning}.\footnote{As an example, IWSLT has a $\approx80$ sequence length with a window size of 4 (Table~\ref{tab:dataset-stats}, appendix).} %

\section{Scalable Document-Level Translation}

\begin{figure}[t]
\centering
\includegraphics[width=0.46\textwidth]{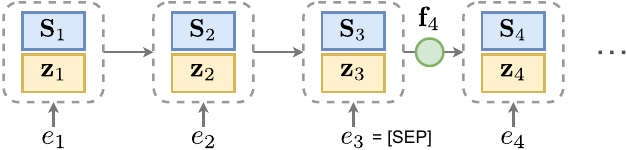}
\caption{\label{fig:gating} Our sentential gating mechanism. $e_1$ and $e_4$ are at the beginnings of two sentences.
}
\end{figure}

We test RFA %
as a linear time and space model to improve the efficiency of document translation. %
We also augment it with a sentential gate to circumvent capacity constraints that come with a long context by injecting a recency inductive bias.
\subsection{Random Feature Attention}
RFA approximates the softmax attention $\operatorname{attn}\left(\vq_t,\{\vk_i\},\{\vv_i\}\right)$ in linear time and space:
\begin{align}
    \textsc{Rfa}\left(\vq_t,\{\vk_i\},\{\vv_i\}\right)=\frac{\boldsymbol{\phi}\left(\vq_t\right) \cdot \mS_t}
    {\boldsymbol{\phi}\left(\vq_t\right) \cdot \vz_t}. \nonumber
\end{align} 
\rev{The randomized nonlinear transformation $\vphi(\cdot)$
is constructed so that $\exp(\vx \cdot \vy)\approx\vphi(\vx)\cdot\vphi(\vy)$~\cite{rahimi2009rff}.} $\mS$, $\vz$ summarize keys and values. We use RFA in cross and causal attention, which are the most impactful for speed and memory, so $\vq_t$ is always from the target sequence. 
In cross attention, $\mS$ and $\vz$ represent the source sequence and are constant for all query positions $t$: $\mS_t = \sum_{i=1}^{N} \boldsymbol{\phi}(\vk_i)\vv_i^\top$ and $\vz_t = \sum_{i=1}^{N} \boldsymbol{\phi}(\vk_i)$.
In causal attention, they represent the target prefix ${i\le t}$: $\mS_t = \sum_{i=1}^{t} \boldsymbol{\phi}(\vk_i)\vv_i^\top = \mS_{t-1} + \boldsymbol{\phi}\left(\vk_t\right)\vv_t^\top$ and $\vz_t = \sum_{i=1}^{t} \boldsymbol{\phi}(\vk_i) = \vz_{t-1} + \boldsymbol{\phi}\left(\vk_t\right)$. 
These recurrent computations are analogous to an RNN with $\mS_t$ and $\vz_t$ as hidden states at step $t$ and enable constant computation per timestep.
RFA serves as a drop-in replacement for attention in transformers.
The encoder and other modules, e.g., feed-forward layers, remain the same.
We refer the reader to \citet{peng2021rfa} for more details on RFA.

\subsection{Sentential Gating} \label{sec:gating}
\citet{schlag2021linear} noted, under the lens of fast weight programmers~\cite{Schmidhuber:91fastweights,schmidhuber1992learning,schmidhuber1993reducing}, that accumulating memory in a purely additive manner, \rev{like $\mS$ and $\vz$} above, will reach a capacity limitation with sequences longer than the size of $\vphi$. %
This is particularly an issue in document-level translation due to the long sequences.

Nevertheless, document translation admits a natural solution to this constraint: inspired by gated RNNs~\citep[\emph{i.a.}]{cho-etal-2014-learning}, we augment RFA with a sentence-level gate to enable dynamic control of
contextual information from the current and previous sentences, and to allow the model to selectively forget about the history to circumvent the capacity constraint. This is illustrated in Figure~\ref{fig:gating}.
For the $t$th word with representation $\ve_t$, we compute a forget gate using the sentence separator token:
\begin{align}
\begin{split}
    f_t &= 
    \begin{cases}
        \sigma( \vw_f\cdot \ve_{t-1}+ b_f) & \text{if }x_t\text{ starts a sentence}\\
        1 & \text{otherwise}\\
    \end{cases}\\
    \mS_t &= f_t \; \mS_{t-1} + \boldsymbol{\phi}\left(\vk_t\right)\vv_t^\top \\
    \vz_t &= f_t \; \vz_{t-1} + \boldsymbol{\phi}\left(\vk_t\right)  \nonumber
\end{split}
\end{align}
$\sigma$ is the sigmoid function; $\vw_f$ and $b_f$ are learned parameters.
The context is decayed when a new sentence starts, and the decay coefficient is reused for all tokens in the same sentence.
Specifically, each sentence $j$ assigns a weight $0 < \prod_{i=\textsc{start}(j')+1}^{\textsc{start}(j)} f_i < 1$ when attending to a previous sentence $j'$, where $\textsc{start}(\cdot)$ is the first token in a sentence.
This introduces an inductive bias that, intuitively, previous sentences are less important in translation, and their representations are decayed.

\paragraph{Relation to Prior Work.} While gating is common in RNNs, it is less clear how it applies to transformers.
\citet{miculicich-etal-2018-document} gated at the sentence level, though hierarchically, while we gate recurrently. 
Ours also contrasts with the per-token gating of \citet{peng2021rfa} which they found ineffective for machine translation.
These two works also take a weighted average of the previous and current sentences while we only decay the former. We show our variant performs better in \S\ref{sec:results}.
\citet{schlag2021linear} used a gate that explicitly models memory removal, but also at the token level.

\section{Experimental Setup} \label{sec:setup}

\paragraph{Datasets and Evaluation.} We experiment with the IWSLT 2015 Chinese-to-English (zh-en) dataset~\cite{Cettolo2015TheI2} with multilingual TED talk captions and the OpenSubtitles2018 English-to-Russian (en-ru) dataset~\cite{lison-etal-2018-opensubtitles2018} with movie and TV subtitles.
We measure document-level BLEU~\cite{papineni-etal-2002-bleu} with SacreBLEU~\cite{post-2018-call}.\footnote{We use fairseq's default setting which has hash \texttt{case.mixed+numrefs.?+smooth.exp+tok.none\\+version.1.5.0} with standalone 13a-tokenization.}
To quantify discourse consistency,
we also use the test sets by \citet{voita-etal-2019-good} based on OpenSubtitles.
We introduce these datasets and their statistics in more detail in \S\ref{sec:dataset-and-processing-details}.

\paragraph{Data Processing.} %
We process each document with a stride-one sliding window of $L$ sentences to obtain our training set.
Following \citet{voita-etal-2019-good} and \citet{ma2021comparison}, we experiment with $L=1$, the sentence-level baseline, and $L=4$. During inference, we use the last translated sentence in each window for evaluation. %
For a more granular analysis, we consider $L\in \{1,2,3,4\}$ for consistency experiments.
More details are in \S\ref{sec:dataset-and-processing-details}.

\paragraph{Model Settings.} We compare RFA and transformer with the concatenation model. For RFA, we experiment with no gating (\textbf{RFA}) and sentential gating (\textbf{RFA-sgate}). To compare our decaying gate choice with prior work (\S\ref{sec:gating}), we run a sentential-gated RFA that takes a weighted average of previous and current text (\textbf{RFA-sgate-avg}). We mostly default to fairseq hyperparameters~\cite{ott2019fairseq}, most suitable for the $L=1$ transformer (\S\ref{sec:hyperparameters}).

\section{Results} \label{sec:results}

\begin{table}[t!]
    \centering
    \aboverulesep=0ex
    \belowrulesep=0ex
    \renewcommand{\arraystretch}{1.2}
    \setlength{\tabcolsep}{6.5pt}
    \begin{tabular}{@{} l @{\hspace{3pt}} c@{\hspace{5pt}}c @{\hspace{5pt}} c@{\hspace{5pt}}c @{}}
        \toprule
        & \multicolumn{2}{c}{\textbf{IWSLT}} & \multicolumn{2}{c}{\textbf{Subtitles}} \\
        \cmidrule(lr){2-3} \cmidrule(lr){4-5}
        \textbf{Window Size $L$} & \textbf{1} & \textbf{4} & \textbf{1} & \textbf{4} \\
        \midrule
        \midrule
        Transformer & 31.7 & 30.4 & 32.6 & 33.1 \\
        \midrule
        RFA & 31.0 & \textbf{30.7} & \textbf{32.9} & \textbf{33.2} \\
        RFA-sgate-avg & --- & \textbf{30.8} & --- & 33.0 \\
        RFA-sgate & --- & \textbf{31.2} & --- & \textbf{33.2} \\
        \bottomrule
    \end{tabular}
    \caption{\label{tab:bleu} BLEU on IWSLT and OpenSubtitles test sets. Bold scores outperform the transformer. %
    }
\end{table}

\begin{figure}[t]
\centering
\includegraphics[width=0.43\textwidth]{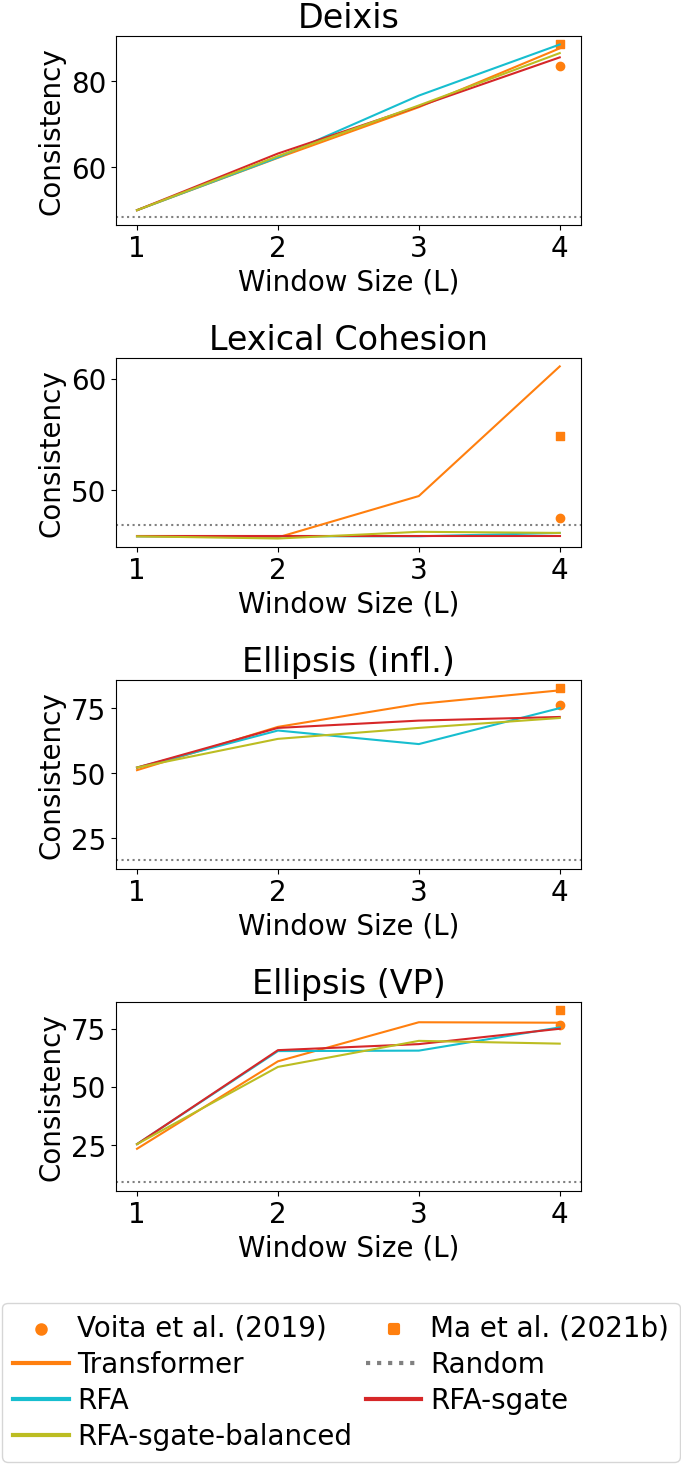}
\caption{\label{fig:consistency} Model performance on the consistency test set, broken down into phenomena. Transformer and RFA are tested with window sizes from 1 to 4. We compare with the baselines in \citet{voita-etal-2019-good} and \citet{ma2021comparison} corresponding to our Transformer $L=4$.
}
\end{figure}

\begin{figure}[t]
\centering
    \includegraphics[width=0.45\textwidth]{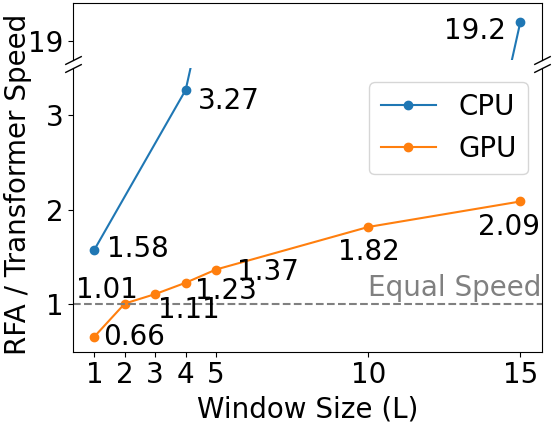}
\caption{\label{fig:speedup} \rev{RFA's inference speedup over the transformer in the number of decoded tokens per second. Each sentence has $\approx20$ tokens (Table~\ref{tab:dataset-stats}, appendix).}
}
\end{figure}

\paragraph{BLEU Scores.} Table~\ref{tab:bleu} shows BLEU scores on IWSLT and OpenSubtitles. 
The sentence-level transformer has the highest IWSLT BLEU, possibly due to defaulting to fairseq hyperparameters optimized for this setting.
With $L=4$, RFA performs slightly better than the transformer, showing suitability for long-text translation. %
Gated RFA further brings a 0.5 BLEU improvement, confirming its utility, but gating has no effect on OpenSubtitles. We hypothesize that with only $\approx$10 tokens per sentence, half of the average length of IWSLT (Table~\ref{tab:dataset-stats}, appendix), \rev{the capacity constraint~\cite{schlag2021linear} is less severe and thus gating would be less useful}.
Our gate also outperforms the averaging variant in \citet{miculicich-etal-2018-document} and \citet{peng2021rfa}, validating its effect on document translation.
Aligning with prior findings \cite{voita-etal-2019-good,ma2021comparison}, longer contexts do not clearly lead to better BLEU, though it improves consistency metrics, to which we turn next.

\paragraph{Discourse Consistency Scores.}

Figure~\ref{fig:consistency} plots the consistency scores in four phenomena for RFA, \rev{including our gated variants}, and the transformer baseline from \citet{voita-etal-2019-good} and \citet{ma2021comparison}. %
We also re-implement this transformer model to control confounding factors in implementation details and to extrapolate to $L<4$, which they did not thoroughly explore.
We compare to a baseline \rev{that chooses its prediction randomly from candidate translations; see \S\ref{sec:dataset-and-processing-details} for details.}

Translating with longer contexts almost always yields better consistency, which is also a setting where RFA achieves better speedup, shown later.
\rev{Gating does not have a clear benefit, aligning with OpenSubtitles' BLEU pattern.}
RFA slightly underperforms the transformer on ellipsis.
We hypothesize that the direct query-key interaction in softmax attention is more suitable for precise long-distance information extraction, sometimes required for consistency, than the RFA approximation. 
\rev{
On lexical cohesion, transformer shows a large variance: with the same architecture and size,
\citet{ma2021comparison}, \citet{voita-etal-2019-good}, and our implementation of $L=4$ transformer
achieve drastically different scores.
\citet{voita-etal-2019-good}'s implementation and RFA fail to outperform the random baseline on this phenomenon.
}
\rev{Reliable evaluation of lexical cohesion, and the related task of word sense disambiguation, are known to be challenging in document translation: models tend to rely on dataset artifacts but not the context, and the attention of well-performing models poorly aligns with the ground-truth required for disambiguation~\citep{kim-etal-2019-document,emelin-etal-2020-detecting,yin-etal-2021-context}.}

\paragraph{Speed.}
\rev{We confirmed the observation from prior work that longer context boosts translation consistency and sometimes BLEU.}
It would be exciting to examine this trend with $L>4$, but to our knowledge, it has little existing evaluation data.
We therefore measure decoding efficiency with a synthetic experiment by decoding for all $L$ with the same trained model.  We focus only on efficiency here, not quality.
We measure the number of decoded tokens per second over the forward pass time on IWSLT's test set.
We follow \citet{ott-etal-2018-scaling} and cache $\vk$ and $\vv$ 
for our \emph{baseline} which substantially increases its speed.
More details are in \S\ref{sec:benmark-details}.

Figure~\ref{fig:speedup} shows RFA's speedup relative to the transformer. \rev{On GPU, without document context, RFA is slower, likely due to its random matrix overhead. Nevertheless, its speed over the transformer roughly linearly increases, agreeing with the theory, up to $2.09\times$ on our longest tested context $L=15$.
RFA enables an even more substantial speedup on other device types. For sentence-level translation, RFA is in fact faster than the transformer by 1.58$\times$ on CPU, and, as \citet{peng2021rfa} reported, by 1.8--1.9$\times$ on TPU. At $L=15$, its CPU speedup increases to 19.2$\times$. Therefore, depending on the use case, such as when decoding on edge devices, RFA could be even more favorable. Furthermore, we used the same batch size between RFA and the transformer. With lower memory complexity, RFA accommodates a larger batch size and in practice achieve a more significant speedup. For example, at $L=15$ on GPU, we found that RFA allows a 5$\times$ batch size and enables a more than $7\times$ speedup.}

\rev{RFA's superior speed makes it an attractive choice to leverage very long contexts. Nevertheless, we are merely extrapolating the utility of long context from our experiments. The extent to which it really helps needs to be verified by future curated test sets. We hope the demonstration of our model's ability to efficiently and effectively process document context could catalyze such efforts.}

\section{Conclusion}

We explored the effectiveness of random feature attention on document translation. Our model substantially improves its speed over a transformer with similar or improved BLEU. Our sentential gate also proves effective, especially on long sequences. While our model may potentially be used to produce toxic or fake information, it also enables more efficient detectors toward such content.

\section*{Limitations}

Limited by existing document translation datasets where ``documents'' are usually relatively short multi-sentence windows, we adopted a semi-synthetic setup for our speed benchmark experiments to examine RFA's effectiveness on long sequences. We believe our results should transfer to real data since decoding speed is mostly a function of sentence length, but this is not a guarantee. Additionally, while RFA would enjoy a better speedup on TPUs as reported in the original RFA paper, we did not have the necessary resources to run experiments on TPUs, so our setup does not fully leverage RFA's potential.

\section*{Acknowledgments}

This work was supported in part by NSF grant 2113530.  Nikolaos Pappas was supported by the Swiss National Science Foundation grant P400P2\_183911.

\bibliography{anthology,custom}

\begin{thebibliography}{42}
\expandafter\ifx\csname natexlab\endcsname\relax\def\natexlab#1{#1}\fi

\bibitem[{Bahdanau et~al.(2015)Bahdanau, Cho, and Bengio}]{Bahdanau2015}
Dzmitry Bahdanau, Kyunghyun Cho, and Yoshua Bengio. 2015.
\newblock \href {https://arxiv.org/pdf/1409.0473.pdf} {Neural machine
  translation by jointly learning to align and translate}.
\newblock In \emph{Proc. of ICLR}.

\bibitem[{Bawden et~al.(2018)Bawden, Sennrich, Birch, and
  Haddow}]{bawden-etal-2018-evaluating}
Rachel Bawden, Rico Sennrich, Alexandra Birch, and Barry Haddow. 2018.
\newblock \href {https://www.aclweb.org/anthology/N18-1118} {Evaluating
  discourse phenomena in neural machine translation}.
\newblock In \emph{Proc. of NAACL}.

\bibitem[{Cettolo et~al.(2015)Cettolo, Niehues, St{\"u}ker, Bentivogli,
  Cattoni, and Federico}]{Cettolo2015TheI2}
M.~Cettolo, J.~Niehues, S.~St{\"u}ker, L.~Bentivogli, R.~Cattoni, and Marcello
  Federico. 2015.
\newblock The {IWSLT} 2015 evaluation campaign.
\newblock In \emph{Proc. of IWSLT}.
\newblock Downloaded from \url{https://wit3.fbk.eu/2015-01}.

\bibitem[{Chen et~al.(2018)Chen, Firat, Bapna, Johnson, Macherey, Foster,
  Jones, Schuster, Shazeer, Parmar, Vaswani, Uszkoreit, Kaiser, Chen, Wu, and
  Hughes}]{chen-etal-2018-best}
Mia~Xu Chen, Orhan Firat, Ankur Bapna, Melvin Johnson, Wolfgang Macherey,
  George Foster, Llion Jones, Mike Schuster, Noam Shazeer, Niki Parmar, Ashish
  Vaswani, Jakob Uszkoreit, Lukasz Kaiser, Zhifeng Chen, Yonghui Wu, and
  Macduff Hughes. 2018.
\newblock \href {https://www.aclweb.org/anthology/P18-1008} {The best of both
  worlds: Combining recent advances in neural machine translation}.
\newblock In \emph{Proc. of ACL}.

\bibitem[{Child et~al.(2019)Child, Gray, Radford, and Sutskever}]{child19}
Rewon Child, Scott Gray, Alec Radford, and Ilya Sutskever. 2019.
\newblock \href {http://arxiv.org/abs/1904.10509} {Generating long sequences
  with sparse transformers}.

\bibitem[{Cho et~al.(2014)Cho, van Merri{\"e}nboer, Gulcehre, Bahdanau,
  Bougares, Schwenk, and Bengio}]{cho-etal-2014-learning}
Kyunghyun Cho, Bart van Merri{\"e}nboer, Caglar Gulcehre, Dzmitry Bahdanau,
  Fethi Bougares, Holger Schwenk, and Yoshua Bengio. 2014.
\newblock \href {https://aclanthology.org/D14-1179} {Learning phrase
  representations using {RNN} encoder{--}decoder for statistical machine
  translation}.
\newblock In \emph{Proc. of EMNLP}.

\bibitem[{Choromanski et~al.(2021)Choromanski, Likhosherstov, Dohan, Song,
  Gane, Sarlos, Hawkins, Davis, Mohiuddin, Kaiser, Belanger, Colwell, and
  Weller}]{choromanski2021rethinking}
Krzysztof~Marcin Choromanski, Valerii Likhosherstov, David Dohan, Xingyou Song,
  Andreea Gane, Tamas Sarlos, Peter Hawkins, Jared~Quincy Davis, Afroz
  Mohiuddin, Lukasz Kaiser, David~Benjamin Belanger, Lucy~J Colwell, and Adrian
  Weller. 2021.
\newblock \href {https://openreview.net/forum?id=Ua6zuk0WRH} {Rethinking
  attention with performers}.
\newblock In \emph{Proc. of ICLR}.

\bibitem[{Donato et~al.(2021)Donato, Yu, and Dyer}]{donato-etal-2021-diverse}
Domenic Donato, Lei Yu, and Chris Dyer. 2021.
\newblock \href {https://aclanthology.org/2021.acl-long.104} {Diverse
  pretrained context encodings improve document translation}.
\newblock In \emph{Proc. of ACL}.

\bibitem[{Emelin et~al.(2020)Emelin, Titov, and
  Sennrich}]{emelin-etal-2020-detecting}
Denis Emelin, Ivan Titov, and Rico Sennrich. 2020.
\newblock \href {https://aclanthology.org/2020.emnlp-main.616} {Detecting word
  sense disambiguation biases in machine translation for model-agnostic
  adversarial attacks}.
\newblock In \emph{Proc. of EMNLP}.

\bibitem[{Guo et~al.(2019)Guo, Qiu, Liu, Shao, Xue, and
  Zhang}]{guo-etal-2019-star}
Qipeng Guo, Xipeng Qiu, Pengfei Liu, Yunfan Shao, Xiangyang Xue, and Zheng
  Zhang. 2019.
\newblock \href {https://www.aclweb.org/anthology/N19-1133} {Star-transformer}.
\newblock In \emph{Proc. of NAACL}.

\bibitem[{Junczys-Dowmunt(2019)}]{junczysdowmunt2019microsoft}
Marcin Junczys-Dowmunt. 2019.
\newblock \href {http://arxiv.org/abs/1907.06170} {Microsoft translator at
  {WMT} 2019: Towards large-scale document-level neural machine translation}.

\bibitem[{Kasai et~al.(2021)Kasai, Peng, Zhang, Yogatama, Ilharco, Pappas, Mao,
  Chen, and Smith}]{kasai2021finetuning}
Jungo Kasai, Hao Peng, Yizhe Zhang, Dani Yogatama, Gabriel Ilharco, Nikolaos
  Pappas, Yi~Mao, Weizhu Chen, and Noah~A. Smith. 2021.
\newblock \href {https://aclanthology.org/2021.emnlp-main.830} {Finetuning
  pretrained transformers into {RNN}s}.
\newblock In \emph{Proc. of EMNLP}.

\bibitem[{Kim et~al.(2019)Kim, Tran, and Ney}]{kim-etal-2019-document}
Yunsu Kim, Duc~Thanh Tran, and Hermann Ney. 2019.
\newblock \href {https://aclanthology.org/D19-6503} {When and why is
  document-level context useful in neural machine translation?}
\newblock In \emph{Proc. of the Fourth Workshop on Discourse in Machine
  Translation}.

\bibitem[{Kingma and Ba(2015)}]{kingma2017adam}
Diederik~P. Kingma and Jimmy Ba. 2015.
\newblock \href {http://arxiv.org/abs/1412.6980} {Adam: {A} method for
  stochastic optimization}.
\newblock In \emph{Proc. of ICLR}.

\bibitem[{Kitaev et~al.(2020)Kitaev, Kaiser, and Levskaya}]{Kitaev2020Reformer}
Nikita Kitaev, Lukasz Kaiser, and Anselm Levskaya. 2020.
\newblock \href {https://openreview.net/forum?id=rkgNKkHtvB} {Reformer: The
  efficient transformer}.
\newblock In \emph{Proc. of ICLR}.

\bibitem[{Koehn et~al.(2007)Koehn, Hoang, Birch, Callison-Burch, Federico,
  Bertoldi, Cowan, Shen, Moran, Zens, Dyer, Bojar, Constantin, and
  Herbst}]{koehn-etal-2007-moses}
Philipp Koehn, Hieu Hoang, Alexandra Birch, Chris Callison-Burch, Marcello
  Federico, Nicola Bertoldi, Brooke Cowan, Wade Shen, Christine Moran, Richard
  Zens, Chris Dyer, Ond{\v{r}}ej Bojar, Alexandra Constantin, and Evan Herbst.
  2007.
\newblock \href {https://www.aclweb.org/anthology/P07-2045} {{M}oses: Open
  source toolkit for statistical machine translation}.
\newblock In \emph{Proc. of ACL}.

\bibitem[{Lison et~al.(2018)Lison, Tiedemann, and
  Kouylekov}]{lison-etal-2018-opensubtitles2018}
Pierre Lison, J{\"o}rg Tiedemann, and Milen Kouylekov. 2018.
\newblock \href {https://www.aclweb.org/anthology/L18-1275}
  {{O}pen{S}ubtitles2018: Statistical rescoring of sentence alignments in
  large, noisy parallel corpora}.
\newblock In \emph{Proc. of LREC}.
\newblock Downloaded the processed version from
  \url{https://github.com/lena-voita/good-translation-wrong-in-context#cadec-data}.

\bibitem[{Lopes et~al.(2020)Lopes, Farajian, Bawden, Zhang, and
  Martins}]{lopes-etal-2020-document}
Ant{\'o}nio Lopes, M.~Amin Farajian, Rachel Bawden, Michael Zhang, and
  Andr{\'e} F.~T. Martins. 2020.
\newblock \href {https://www.aclweb.org/anthology/2020.eamt-1.24}
  {Document-level neural {MT}: A systematic comparison}.
\newblock In \emph{Proc. of EAMT}.

\bibitem[{Ma et~al.(2021{\natexlab{a}})Ma, Kong, Wang, Zhou, May, Ma, and
  Zettlemoyer}]{ma2021luna}
Xuezhe Ma, Xiang Kong, Sinong Wang, Chunting Zhou, Jonathan May, Hao Ma, and
  Luke Zettlemoyer. 2021{\natexlab{a}}.
\newblock \href {https://openreview.net/forum?id=GWRkOYr4jxQ} {Luna: Linear
  unified nested attention}.
\newblock In \emph{Proc. of NeurIPS}.

\bibitem[{Ma et~al.(2021{\natexlab{b}})Ma, Edunov, and Auli}]{ma2021comparison}
Zhiyi Ma, Sergey Edunov, and Michael Auli. 2021{\natexlab{b}}.
\newblock \href {http://arxiv.org/abs/2101.11040} {A comparison of approaches
  to document-level machine translation}.

\bibitem[{Maruf et~al.(2021)Maruf, Saleh, and Haffari}]{maruf2021survey}
Sameen Maruf, Fahimeh Saleh, and Gholamreza Haffari. 2021.
\newblock \href {https://doi.org/10.1145/3441691} {A survey on document-level
  neural machine translation: Methods and evaluation}.
\newblock \emph{ACM Comput. Surv.}, 54(2).

\bibitem[{Miculicich et~al.(2018)Miculicich, Ram, Pappas, and
  Henderson}]{miculicich-etal-2018-document}
Lesly Miculicich, Dhananjay Ram, Nikolaos Pappas, and James Henderson. 2018.
\newblock \href {https://www.aclweb.org/anthology/D18-1325} {Document-level
  neural machine translation with hierarchical attention networks}.
\newblock In \emph{Proc. of EMNLP}.

\bibitem[{M{\"u}ller et~al.(2018)M{\"u}ller, Rios, Voita, and
  Sennrich}]{muller-etal-2018-large}
Mathias M{\"u}ller, Annette Rios, Elena Voita, and Rico Sennrich. 2018.
\newblock \href {https://www.aclweb.org/anthology/W18-6307} {A large-scale test
  set for the evaluation of context-aware pronoun translation in neural machine
  translation}.
\newblock In \emph{Proc. of WMT}.

\bibitem[{Ott et~al.(2019)Ott, Edunov, Baevski, Fan, Gross, Ng, Grangier, and
  Auli}]{ott2019fairseq}
Myle Ott, Sergey Edunov, Alexei Baevski, Angela Fan, Sam Gross, Nathan Ng,
  David Grangier, and Michael Auli. 2019.
\newblock fairseq: A fast, extensible toolkit for sequence modeling.
\newblock In \emph{Proc. of NAACL}.

\bibitem[{Ott et~al.(2018)Ott, Edunov, Grangier, and
  Auli}]{ott-etal-2018-scaling}
Myle Ott, Sergey Edunov, David Grangier, and Michael Auli. 2018.
\newblock \href {https://www.aclweb.org/anthology/W18-6301} {Scaling neural
  machine translation}.
\newblock In \emph{Proc. of WMT}.

\bibitem[{Papineni et~al.(2002)Papineni, Roukos, Ward, and
  Zhu}]{papineni-etal-2002-bleu}
Kishore Papineni, Salim Roukos, Todd Ward, and Wei-Jing Zhu. 2002.
\newblock \href {https://www.aclweb.org/anthology/P02-1040} {{B}leu: a method
  for automatic evaluation of machine translation}.
\newblock In \emph{Proc. of ACL}.

\bibitem[{Peng et~al.(2021)Peng, Pappas, Yogatama, Schwartz, Smith, and
  Kong}]{peng2021rfa}
Hao Peng, Nikolaos Pappas, Dani Yogatama, Roy Schwartz, Noah Smith, and
  Lingpeng Kong. 2021.
\newblock Random feature attention.
\newblock In \emph{Proc. of ICLR}.

\bibitem[{Post(2018)}]{post-2018-call}
Matt Post. 2018.
\newblock \href {https://www.aclweb.org/anthology/W18-6319} {A call for clarity
  in reporting {BLEU} scores}.
\newblock In \emph{Proc. of WMT}.
\newblock Evaluation script at \url{https://github.com/mjpost/sacrebleu}.

\bibitem[{Rahimi and Recht(2008)}]{rahimi2009rff}
Ali Rahimi and Benjamin Recht. 2008.
\newblock \href
  {https://proceedings.neurips.cc/paper/2007/file/013a006f03dbc5392effeb8f18fda755-Paper.pdf}
  {Random features for large-scale kernel machines}.
\newblock In \emph{Proc. of NeurIPS}.

\bibitem[{Schlag et~al.(2021)Schlag, Irie, and Schmidhuber}]{schlag2021linear}
Imanol Schlag, Kazuki Irie, and J\"urgen Schmidhuber. 2021.
\newblock Linear transformers are secretly fast weight programmers.
\newblock In \emph{Proc. of ICML}.

\bibitem[{Schmidhuber(1991)}]{Schmidhuber:91fastweights}
J{\"u}rgen Schmidhuber. 1991.
\newblock Learning to control fast-weight memories: An alternative to recurrent
  nets.
\newblock Technical Report FKI-147-91, Institut f\"{u}r Informatik, Technische
  Universit\"{a}t M\"{u}nchen.

\bibitem[{Schmidhuber(1992)}]{schmidhuber1992learning}
J{\"u}rgen Schmidhuber. 1992.
\newblock Learning to control fast-weight memories: An alternative to dynamic
  recurrent networks.
\newblock \emph{Neural Computation}, 4(1):131--139.

\bibitem[{Schmidhuber(1993)}]{schmidhuber1993reducing}
J{\"u}rgen Schmidhuber. 1993.
\newblock Reducing the ratio between learning complexity and number of time
  varying variables in fully recurrent nets.
\newblock In \emph{Proc. of ICANN}.

\bibitem[{Sennrich et~al.(2016)Sennrich, Haddow, and
  Birch}]{sennrich-etal-2016-neural}
Rico Sennrich, Barry Haddow, and Alexandra Birch. 2016.
\newblock \href {https://www.aclweb.org/anthology/P16-1162} {Neural machine
  translation of rare words with subword units}.
\newblock In \emph{Proc. of ACL}.

\bibitem[{Tiedemann and Scherrer(2017)}]{tiedemann-scherrer-2017-neural}
J{\"o}rg Tiedemann and Yves Scherrer. 2017.
\newblock \href {https://www.aclweb.org/anthology/W17-4811} {Neural machine
  translation with extended context}.
\newblock In \emph{Proc. of the Third Workshop on Discourse in Machine
  Translation}.

\bibitem[{Tu et~al.(2018)Tu, Liu, Shi, and Zhang}]{tu2017learning}
Zhaopeng Tu, Yang Liu, Shuming Shi, and Tong Zhang. 2018.
\newblock \href {https://doi.org/10.1162/tacl_a_00029} {{Learning to Remember
  Translation History with a Continuous Cache}}.
\newblock \emph{Transactions of the Association for Computational Linguistics},
  6:407--420.

\bibitem[{Vaswani et~al.(2017)Vaswani, Shazeer, Parmar, Uszkoreit, Jones,
  Gomez, Kaiser, and Polosukhin}]{vaswani2017}
Ashish Vaswani, Noam Shazeer, Niki Parmar, Jakob Uszkoreit, Llion Jones,
  Aidan~N Gomez, \L~ukasz Kaiser, and Illia Polosukhin. 2017.
\newblock \href
  {https://proceedings.neurips.cc/paper/2017/file/3f5ee243547dee91fbd053c1c4a845aa-Paper.pdf}
  {Attention is all you need}.
\newblock In \emph{Proc. of NeurIPS}.

\bibitem[{Voita et~al.(2019)Voita, Sennrich, and Titov}]{voita-etal-2019-good}
Elena Voita, Rico Sennrich, and Ivan Titov. 2019.
\newblock \href {https://www.aclweb.org/anthology/P19-1116} {When a good
  translation is wrong in context: Context-aware machine translation improves
  on deixis, ellipsis, and lexical cohesion}.
\newblock In \emph{Proc. of ACL}.
\newblock Dataset and scoring script at
  \url{https://github.com/lena-voita/good-translation-wrong-in-context}.

\bibitem[{Wang et~al.(2019)Wang, Li, Xiao, Zhu, Li, Wong, and
  Chao}]{wang-etal-2019-learning}
Qiang Wang, Bei Li, Tong Xiao, Jingbo Zhu, Changliang Li, Derek~F. Wong, and
  Lidia~S. Chao. 2019.
\newblock \href {https://www.aclweb.org/anthology/P19-1176} {Learning deep
  transformer models for machine translation}.
\newblock In \emph{Proc. of ACL}.

\bibitem[{Wang et~al.(2020)Wang, Li, Khabsa, Fang, and Ma}]{wang2020linformer}
Sinong Wang, Belinda~Z. Li, Madian Khabsa, Han Fang, and Hao Ma. 2020.
\newblock \href {http://arxiv.org/abs/2006.04768} {Linformer: Self-attention
  with linear complexity}.

\bibitem[{Yin et~al.(2021)Yin, Fernandes, Pruthi, Chaudhary, Martins, and
  Neubig}]{yin-etal-2021-context}
Kayo Yin, Patrick Fernandes, Danish Pruthi, Aditi Chaudhary, Andr{\'e} F.~T.
  Martins, and Graham Neubig. 2021.
\newblock \href {https://aclanthology.org/2021.acl-long.65} {Do context-aware
  translation models pay the right attention?}
\newblock In \emph{Proc. of ACL}.

\bibitem[{Zhang et~al.(2018)Zhang, Luan, Sun, Zhai, Xu, Zhang, and
  Liu}]{zhang-etal-2018-improving}
Jiacheng Zhang, Huanbo Luan, Maosong Sun, Feifei Zhai, Jingfang Xu, Min Zhang,
  and Yang Liu. 2018.
\newblock \href {https://www.aclweb.org/anthology/D18-1049} {Improving the
  transformer translation model with document-level context}.
\newblock In \emph{Proc. of EMNLP}.

\end{thebibliography}
\bibliographystyle{acl_natbib}

\clearpage
\appendix
\begin{table}[t!]
\centering
\begin{tabular}{l@{\hspace{6pt}}c@{\hspace{6pt}}c@{\hspace{6pt}}r@{\hspace{6pt}}r@{\hspace{6pt}}r@{\hspace{6pt}}r@{\hspace{6pt}}r}
\toprule
\multirow{2}{*}{\textbf{Dataset}} & \multirow{2}{*}{\textbf{Lg.}} & \textbf{Train} & \textbf{Dev.} & \textbf{Test} & \textbf{Sent.} & \textbf{Tok.} \\
&& \textbf{Docs} & \textbf{Docs} & \textbf{Docs} & \textbf{/doc} & \textbf{/sent.} \\
\midrule
\multirow{2}{*}{IWSLT} & zh & \multirow{2}{*}{1713} & \multirow{2}{*}{8} & \multirow{2}{*}{56} & \multirow{2}{*}{121.5} & 20.4 \\
& en &&&&& 22.6 \\
\midrule
\multirow{2}{*}{Sub.} & en & \multirow{2}{*}{1.5M} & \multirow{2}{*}{10K} & \multirow{2}{*}{10K} & \multirow{2}{*}{4} & 10.3 \\
& ru &&&&& 9.5 \\
\midrule
Sub.- & en & \multirow{2}{*}{---} & \multirow{2}{*}{2K} & \multirow{2}{*}{16K} & \multirow{2}{*}{4} & 10.5 \\
Cons. & ru &&&&& 9.6 \\
\bottomrule
\end{tabular}
\caption{\label{tab:dataset-stats} Dataset statistics of IWSLT, OpenSubtitles, and the consistency test sets for OpenSubtitles. We follow \citet{ma2021comparison} in treating the four-sentence windows of OpenSubtitles as separate documents. The number of sentences per document and BPE tokens per sentence are averaged across all splits, except for OpenSubtitles-Consistency, which are only averaged across the development and test sets.
}
\end{table}

\section{Appendix}

\subsection{Dataset and Processing Details} \label{sec:dataset-and-processing-details}

The IWSLT 2015 dataset contains multilingual TED talk captions.
Following \citet{miculicich-etal-2018-document}, we use the Chinese-to-English (zh-en) portion and use the \emph{dev2010} subset for development and \emph{tst2010-2013} for testing. 
We also use the processed OpenSubtitles2018 English-to-Russian (en-ru) dataset by \citet{voita-etal-2019-good}.
The consistency test sets by \citet{voita-etal-2019-good} measure (i) pronominal formality consistency (\textbf{deixis}), (ii) word choice consistency (\textbf{lexical cohesion}), (iii) inflection prediction accuracy of syntactically ambiguous words due to ellipsis (\textbf{ellipsis (inflection)}), and (iv) elided verb prediction accuracy (\textbf{ellipsis (VP)}). Models choose the candidate translation most consistent with the context and are scored with accuracy. \rev{We present an example taken from \citet{voita-etal-2019-good} for lexical cohesion: the source English sequence is ``Not for \underline{Julia}. \underline{Julia} has a taste for taunting her victims.'' and the target Russian translation candidates consist of two sequences, here transcribed with the Latin script: (a) ``Ne dlya \underline{Dzhulii}. \underline{Yuliya} umeyet draznit’ svoikh zhertv.''; and (b) ``Ne dlya \underline{Dzhulii}. \underline{Dzhulii} umeyet draznit’ svoikh zhertv.'' The model is expected to choose (b) as it uses the same consistent translation for the name ``\underline{Julia}.''
Our random baseline randomly picks its translation from the candidate set.}
Table~\ref{tab:dataset-stats} summarizes dataset statistics.

We follow the tokenization of \citet{miculicich-etal-2018-document}. For all datasets, we first tokenize and truecase English and Russian with Moses~\cite{koehn-etal-2007-moses} and tokenize Chinese using Jieba.\footnote{\url{https://github.com/fxsjy/jieba}} We then run byte-pair encoding~\cite{sennrich-etal-2016-neural} on the concatenation of the training sets of the source and target languages using 30k splits, separately done for each dataset.

\subsection{Hyperparameters and Training Details} \label{sec:hyperparameters}

Following \citet{vaswani2017} and \citet{peng2021rfa}, we use 6-layer transformers with 512 hidden dimension and 8 attention heads for both the encoder and decoder. Both RFA and the transformer baselines have 53M trainable parameters for IWSLT and 49M for OpenSubtitles, with the difference caused by different vocabulary sizes. We train all models in mixed-precision. We use the Adam optimizer~\cite{kingma2017adam} with peak learning rate searched in $\{0.0005, 0.001\}$ warmed up through 8000 updates and an effective batch size of 16,384 in the number of tokens. We use beam size 4 for decoding. All other hyperparameters follow the recommendation in fairseq~\cite{ott2019fairseq}.\footnote{\url{https://github.com/pytorch/fairseq/tree/v0.10.0/examples/translation##iwslt14-german-to-english-transformer}} 
For RFA-sgate, to better enforce the inductive bias where sentences further away are less important, we treat the initialization of $b_f$ in the sentential gating equation as a hyperparameter, searched in $\{1,2\}$, instead of setting it to zero as in RFA. We search the RFA cross attention projection dimension $+$ causal attention projection dimension in $\{128+64, 256+32\}$. We only employ gating in causal attention as we found it to hurt the performance when added in cross attention in preliminary experiments, \rev{degrading the performance by around 1 BLEU on IWSLT}. \rev{According to \citet{donato-etal-2021-diverse}, source context is more important than target context, so it is possible that the model benefits from a non-decayed history on the source side.}

We use early stopping with a patience of 10 epochs based on development set performance. %
\citet{voita-etal-2019-good} observed that BLEU and consistency scores exhibit different training dynamics. We, therefore, train separate OpenSubtitles models when measuring BLEU versus consistency and use the respective metric for early stopping. 

We manually tune the hyperparameters mentioned above based on the development set performance with the corresponding metric (i.e., BLEU or consistency). All final models use 0.001 learning rate. The final IWSLT RFA models use $b_f=2$ and RFA projection dimension $256+32$; OpenSubtitles (BLEU) RFA models use $b_f=1$ and RFA projection dimension $256+32$; OpenSubtitles (consistency) RFA models use RFA projection dimension $128+64$.

We perform all training on a single NVIDIA 2080 Ti GPU. \rev{Depending on the dataset, window size, and model variant, each training run takes approximately 3.5 hours to a day.}

\subsection{Speedup Benchmark Details} \label{sec:benmark-details}

\begin{table}[t!]
\centering
\begin{tabular}{c|c@{\hspace{8pt}}c@{\hspace{8pt}}c@{\hspace{8pt}}c@{\hspace{8pt}}c@{\hspace{8pt}}c@{\hspace{8pt}}c}
\toprule
$L$&1&2&3&4&5&10&15\\
\midrule
$B$&1024&512&512&256&256&128&96 \\
\bottomrule
\end{tabular}
\caption{\label{tab:batch-sizes} \rev{The batch size ($B$) used to decode each window size $L$.}
}
\end{table}

\rev{
In our synthetic benchmark setup, we decode under all $L$ with the same trained model which allows us to examine the trend with a larger $L$. We use the smallest contexted model, $L=2$, and verified that using it to decode in $L=4$ yields a similar length distribution as an actual trained $L=4$ model, confirming that this setup accurately reflects long context output length patterns.
We benchmark with ungated RFA.\footnote{The speed difference between the RFA variants is negligible as gating requires minimal additional computation. This is also confirmed by \citet{peng2021rfa}, where their per-token gating has the same speedup as no gating.}
To simulate real-world usage, we use the largest possible batch size for each window size that fits on a single 32GB A100 GPU, the GPU that we use for all benchmark runs. In practice, as the transformer has larger memory consumption and since we use the same batch size between RFA and the transformer, this is the batch size that saturates the transformer. We only search the batch size over $2^k$ and $1.5 \times 2^k$ for integer $k$ for tractability. We report the batch sizes used in Table~\ref{tab:batch-sizes}. The CPU experiments use the same batch sizes.
We conduct this analysis on IWSLT as we believe OpenSubtitles represent a different genre from many settings where long contexts are expected to be useful, though in this synthetic setup, the trend would be similar when the sequence length is controlled. The CPU experiments are run with six 2.2GHz Intel Cascade Lake CPUs.
}

\end{document}